\documentclass[a4paper, 10pt]{report}
\usepackage[Lenny]{fncychap}
\usepackage{multicol}
\usepackage{type1ec}
\usepackage{lettrine}
\usepackage[T1]{fontenc}
\usepackage[latin1]{inputenc}
\usepackage[english]{babel}
\usepackage[tight, english]{minitoc}

\usepackage{lipsum}
\usepackage{epigraph}
\usepackage{booktabs}
\usepackage{graphicx}
\usepackage{array}
\usepackage{multirow}
\usepackage{caption}

\newenvironment{Figure}
  {\par\medskip\noindent\minipage{\linewidth}}
  {\endminipage\par\medskip}
\usepackage{amsmath}
\usepackage{amssymb}
\usepackage{amsfonts}
\usepackage{amsthm}
\usepackage{mathtools}
\usepackage{mathdots}

\renewcommand{\epsilon}{\varepsilon}
\renewcommand{\theta}{\vartheta}
\renewcommand{\rho}{\varrho}
\renewcommand{\phi}{\varphi}

\DeclarePairedDelimiter{\floor}{\lfloor}{\rfloor}
\makeatletter
\renewcommand*\env@matrix[1][*\c@MaxMatrixCols c]{%
  \hskip -\arraycolsep
  \let\@ifnextchar\new@ifnextchar
  \array{#1}}
\makeatother
\makeatletter
\DeclareRobustCommand*{\bfseries}{%
  \not@math@alphabet\bfseries\mathbf
  \fontseries\bfdefault\selectfont
  \boldmath
}
\makeatother
\usepackage[a4paper, top=2.5cm, bottom=3cm, left=1.5cm, right=1.5cm, heightrounded, bindingoffset=5mm,
marginparwidth=3cm, marginparsep=5mm]{geometry}
\usepackage[italian]{varioref}
\usepackage[pdfauthor={Lou Marvin Caraig}, pdftitle={A New Training Algorithm for Kanerva's Sparse Distributed Memory}, pdfmenubar=false]{hyperref}
\let\oldmarginpar\marginpar
  \renewcommand\marginpar[1]
  {\-\oldmarginpar[\raggedleft\footnotesize\ \textit{#1}]%
  {\raggedright\footnotesize \textit{#1}}}

\author{\emph{Lou Marvin Caraig}}
\title{A New Training Algorithm for Kanerva's Sparse Distributed Memory}
\newcommand{\mymail}{$\mathtt{\mbox{loumarvincaraig.unifi@gmail.com}}$}

\begin{document}

		\pagestyle{plain}
		\begin{center}
		\Large{A New Training Algorithm for Kanerva's Sparse Distributed Memory}\\[0.5cm]
		\normalsize{Lou Marvin Caraig}\\
		\normalsize{Department of Systems and Computer Science}\\
		\normalsize{University of Florence}\\
		\normalsize{Email: \mymail}\\[0.5cm]
		\end{center}
		\begin{multicols}{2}{
		\small{\textbf{\emph{Abstract} --- The Sparse Distributed Memory proposed by Pentii Kanerva (SDM in short) was thought to be a model of human long term memory. The architecture of the SDM permits to store binary patterns and to retrieve them using partially matching patterns. However Kanerva's model is especially efficient only in handling random data. The purpose of this article is to introduce a new approach of training Kanerva's SDM that can handle efficiently non-random data, and to provide it the capability to recognize inverted patterns. This approach uses a signal model which is different from the one proposed for different purposes by Hely, Willshaw and Hayes in $[4]$. This article additionally suggests a different way of creating hard locations in the memory despite the Kanerva's static model.}}\\[0.2cm]

		\subsection{Introduction}
		\normalsize
		In $1984$ Pentii Kanerva introduced the Sparse Distributed Memory trying to modelize the human memory, in particular the long term memory. The idea was that different concepts in our minds are as two points in a high-dimensional space, where the distance is higher as the concepts are more different. The SDM consists in a reasonable large number of memory locations (\emph{hard locations}) randomly distributed in throughout the address space $\{0,1\}^n$ where $n$ is the lenght of the patterns. Furthermore at every location there is a vector of counters initialized to $0$ with the same lenght of the patterns (there is a counter for every bit of the address). In Kanerva's paper a pattern can be memorized in a location different from the binary string representing the pattern itself, anyway in this article a pattern to be stored always addresses itself: patterns are \emph{self-addressing}.
		
		The storage of a pattern consists in updating the vector of counters of every locations which are at a lower distance than a selected one called \emph{radius}. With \emph{distance} between two binary strings is intended the \emph{Hamming distance}, which is the number of different bits between those. When a location is whithin the hypersphere centered on the input pattern, the counters of every locations are updated as follows: a $1$ in the input pattern increases by $1$ the value of the counter at the corresponding position in the vector while a $0$ decreases the value by $1$.
		
		The retrieval consists in summing all the vector of counters whose addresses are within the hypersphere centered on the retrieval pattern, and in applying the Kronecker's delta function to every bit of the sum-vector using $0$ as threshold. The retrieval can also be iterated using the previous recalled pattern.
		
\begin{Figure}
\label{fig:originalSDM}
\centering
\includegraphics[width=\columnwidth]{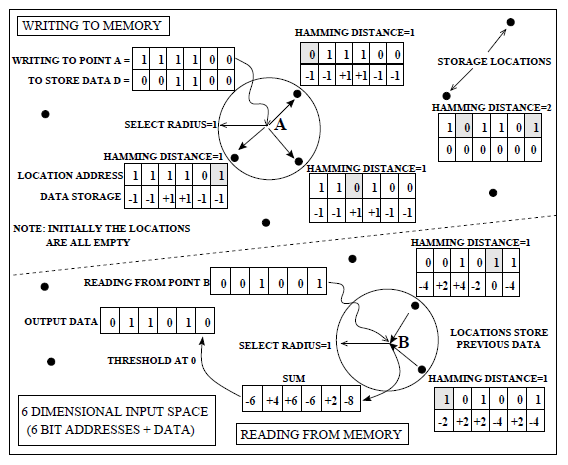}
\captionof{figure}{\small{A SDM with $6$ bit patterns and $\text{radius} = 1$. At the top visualizes the storage algorithm and at bottom visualizes the retrieval algorithm as explained in the introduction.}}
\end{Figure}
		
		The new approaches explained in this article try to overcome the limit of Kanerva's original SDM of handling efficiently only random data. The main reason this problem is important to be solved for is that in the real world random data are rare. These new approaches, that are mathematically translated in different training algorithms, are inspired by thinking about human capability to recognize inverted patterns and about the utilization of both short term and long term memory in recognizing a pattern (an odour, a voice, ect.) depending on the information a person has.
		
		Seeing a black logo on a white background is sufficient for our brain to recognize (most of the times) the same logo, but with inverted colors. That said, why a SDM should not be able to do the same thing? As in~$[4]$ the training algorithm presented hereunder use real values for the counters, and not integers as in Kanerva's model.

\subsection{Maximizing the choice of the hard locations from the address space}
		To use the memory at its best, the new approach do not use the Kanerva's original statical model of creating the hard locations, but the construction of the SDM is dinamic and the locations are created at each storage of a pattern. In details every pattern, before being stored, generates some memory locations in order to maximize their usefulness. The memory locations to create in the SDM are generated by corrupting the input pattern by some given percentage of noise. The percentages of corruption depend on the training algorithm and the number of generated addresses depends on the size of the pattern. In the tests performed by the author, the addresses are generated as follows. $n$ indicates the number of hard locations and $p$ the percentage of corruption:
		\begin{itemize}
		\item $n = 2^{\floor{1.75\% \text{ of the pattern size}}}$, $p = 5\%$
		\item $n = 2^{\floor{1.50\% \text{ of the pattern size}}}$, $p = 10\%$
		\item $n = 2^{\floor{1.25\% \text{ of the pattern size}}}$, $p = 15\%$
		\item $n = 2^{\floor{1.00\% \text{ of the pattern size}}}$, $p = 20\%$
		\item $n = 2^{\floor{0.75\% \text{ of the pattern size}}}$, $p = 25\%$.
		\end{itemize}
		The value of corruption is so that the vectors of counters of all the addresses are updated, hence there are not useless locations. These values can be suitable for Kanerva's SDM, but in the Signal Decay model described hereunder the addresses far from the one identified by the input pattern are also needed, so the number of vectors and the percentage of corruption are modified as follows:
		\begin{itemize}
		\item $n = 2^{\floor{1.50\% \text{ of the pattern size}}}$, $p = 5\%$
		\item $n = 2^{\floor{1.25\% \text{ of the pattern size}}}$, $p = 10\%$
		\item $n = 2^{\floor{1.00\% \text{ of the pattern size}}}$, $p = 15\%$
		\item $n = 2^{\floor{0.75\% \text{ of the pattern size}}}$, $p = 20\%$
		\item $n = 2^{\floor{1.00\% \text{ of the pattern size}}}$, $p = 85\%$
		\item $n = 2^{\floor{1.25\% \text{ of the pattern size}}}$, $p = 90\%$
		\item $n = 2^{\floor{1.50\% \text{ of the pattern size}}}$, $p = 95\%$.
		\end{itemize}
		
		This dinamic way of creating hard locations in the SDM guarantees that every locations are used. These values are just indicative and could be surely chosen better. Anyway the tests demonstrated that these values are already satisfactory comparing the performances of the Kanerva's original SDM despite the new models.

		\subsection{Signal Decay model}
		The differences between the signal model exposed in this article and the signal model exposed in~$[4]$ are:
		\begin{itemize}
		\item the signal in~$[4]$ loses a percentage of strenght at every location reached, while here the strength of the signal is a function of the Hamming distance
		\item the strength of the signal in~$[4]$ does not increase after reaching $0$, while here the minimum strength is at $d = \frac{n}{2}$, where $d$ indicates the hamming distance, and increaseses again heading to $d = n$.
		\end{itemize}
		
		In particular the function used for the strength of the signal is a sine wave:
		\begin{equation}
		\label{eq:signalstrength}
		\text{signal}(d) =
		\begin{cases}
		\frac{1}{2}\left(\cos\left(\frac{2\pi}{n}d\right)+1\right)	& d\in\left[0,\frac{n}{2}\right)\\
		\frac{m}{2}\left(\cos\left(\frac{2\pi}{n}d\right)+1\right)	& d\in\left[\frac{n}{2},n\right]
		\end{cases}
		\end{equation}
		where $m$ is the percentage of the maximum of the lowered signal. In the tests a value of $0.20$ demonstrated that the SDM achieves good performances. The graph of the function is visible in figure~$2$. So instead of increasing or decreasing a counter by $1$, a counter is increased or decreased by $\text{signal}(d)$ where $d$ is the Hamming distance between the input pattern and the address of the location of the corresponding vector of counters.
		
\begin{Figure}
\label{fig:sigdecaygraph}
\centering
\includegraphics[width=\columnwidth]{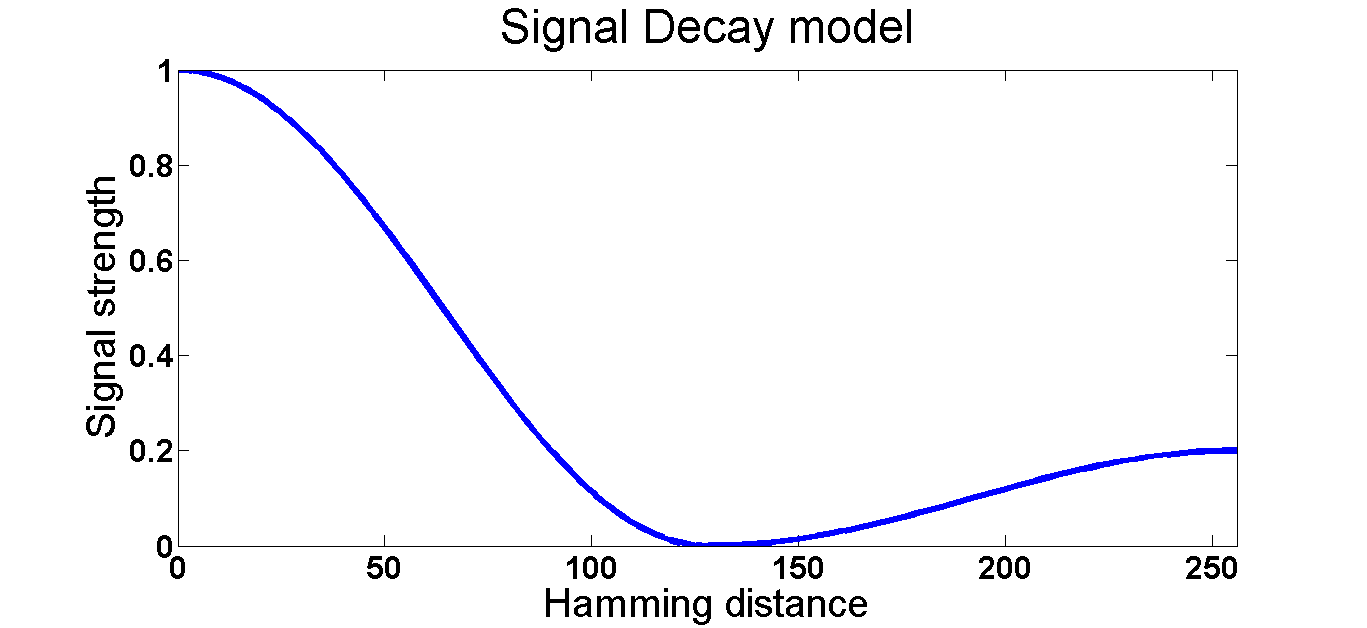}
\captionof{figure}{\small{Graph of the function of the signal strength in the Signal Decay model.}}
\end{Figure}
		
		As in~$[4]$ the SDM is more flexible having no need to select a storage radius. With this Signal Decay model not only the nearest locations are rewarded by a stronger signal, but also the farest ones.
		
		The necessity to generate also far locations as explained in the previous section, is justified by the utilization of the sine wave function described in $(1)$. Not generating those locations will infact not permit the SDM to memorize the informations about the complemented pattern that needs to be memorized far from the input pattern. That said, using for example $20\%$ corrupted patterns of a black 'A' on a white background to train the SDM, will permit the SDM to retrieve the pattern both using a corrupted black 'A' on a white background and using a corrupted white 'A' on a black background (figure~3).
		
		This training algorithm removes the need to select a radius for the storage, but continues to require a radius for the retrieval. Infact the retrieval algorithm is identical to the Kanerva's original SDM, the only difference is that instead of summing integer values are summed real values.
		
\begin{Figure}
\label{fig:testsigdecay}
\centering
\includegraphics[width=\columnwidth]{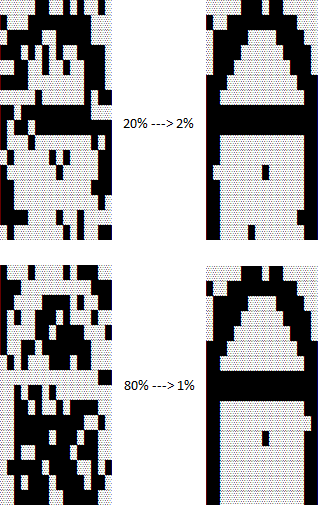}
\captionof{figure}{\small{Capability of a SDM trained with the Signal Decay model to recognize patterns corrupted both by $20\%$ and by $80\%$ even though the training set contained only patterns corrupted by $20\%$.}}
\end{Figure}

		\subsection{Tests}
			The patterns chosen for the tests are visible in figure $4$.
		
		In all the performed tests the pattern size is $256$ and the selected radius is $35\%$ of the pattern size, or $89$. Every tested SDM has been tested with different patterns differing by the Q-factor, where with Q-factor is intended the number of $1$'s in any given pattern. The Q-factor values chosen for the tests are $32$, $64$, $96$, $128$. Every trained SDM has been trained with only a pattern, in particular using $5$ patterns corrupted by $15\%$, $5$ corrupted by $20\%$ and $5$ corrupted by $25\%$.
				
		The tested SDM are:
		\begin{itemize}
		\item Kanerva's original SDM trained with the original algorithm with static creation of hard locations
		\item Signal Decay model with dinamic creation of hard locations
		\end{itemize}
		
\begin{Figure}
\label{fig:patterns}
\centering
\includegraphics[width=\columnwidth]{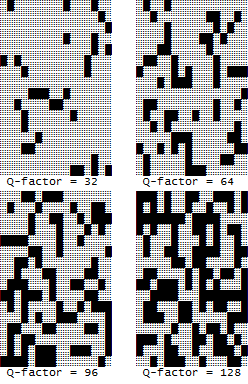}
\captionof{figure}{\small{Patterns used for the tests.}}
\end{Figure}
		
		\subsection{Results}
		Starting to discuss about the results of the tests performed on the straight-trained SDMs using a retrieval pattern with a percentage of corruption between $5\%$ and $30\%$, the new modela have better performances despite Kanerva's original SDM as it is possibile to see in figure~$5$. The difference in performances are moreover visible when the Q-factor is low. Kanerva's SDM cannot even recognize the pattern which Q-factor is $128$ using a corrupted input by $30\%$ (figure~$6$). It is also important to emphasize that the results here exposed refer to the number of bit-errors of the first self-addressing pattern recalled during the retrieval. The reason behind this choice is that during a real retrieval (not a test as in this case) the real pattern is unknown for most of the times, and it can be convenient to stop the retrieval iteration when a self-addressing pattern is recalled.
		
\begin{Figure}
\label{fig:test1}
\centering
\includegraphics[width=\columnwidth]{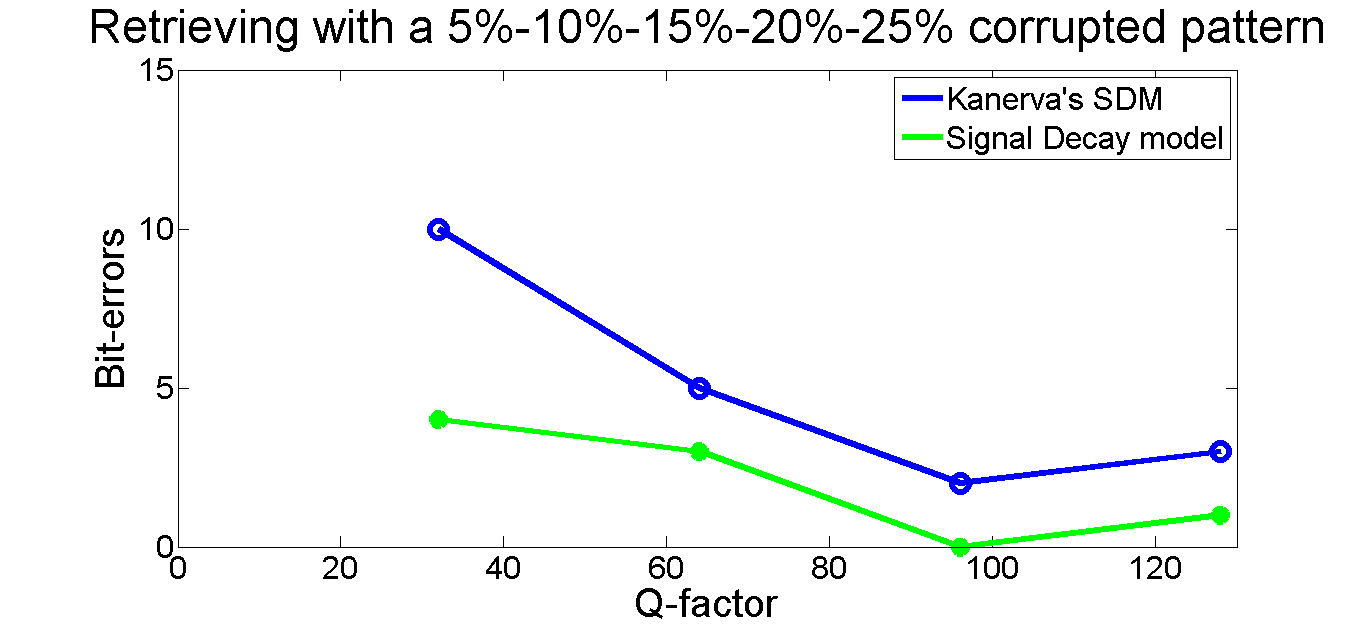}
\captionof{figure}{\small{Results of the test performed on a SDM using a retrieval pattern with a percentage of corruption between $5\%$ and $25\%$. Visualizes the better performances achieved by the new model despite the Kanerva's SDM.}}
\end{Figure}
\begin{Figure}
\label{fig:test2}
\centering
\includegraphics[width=\columnwidth]{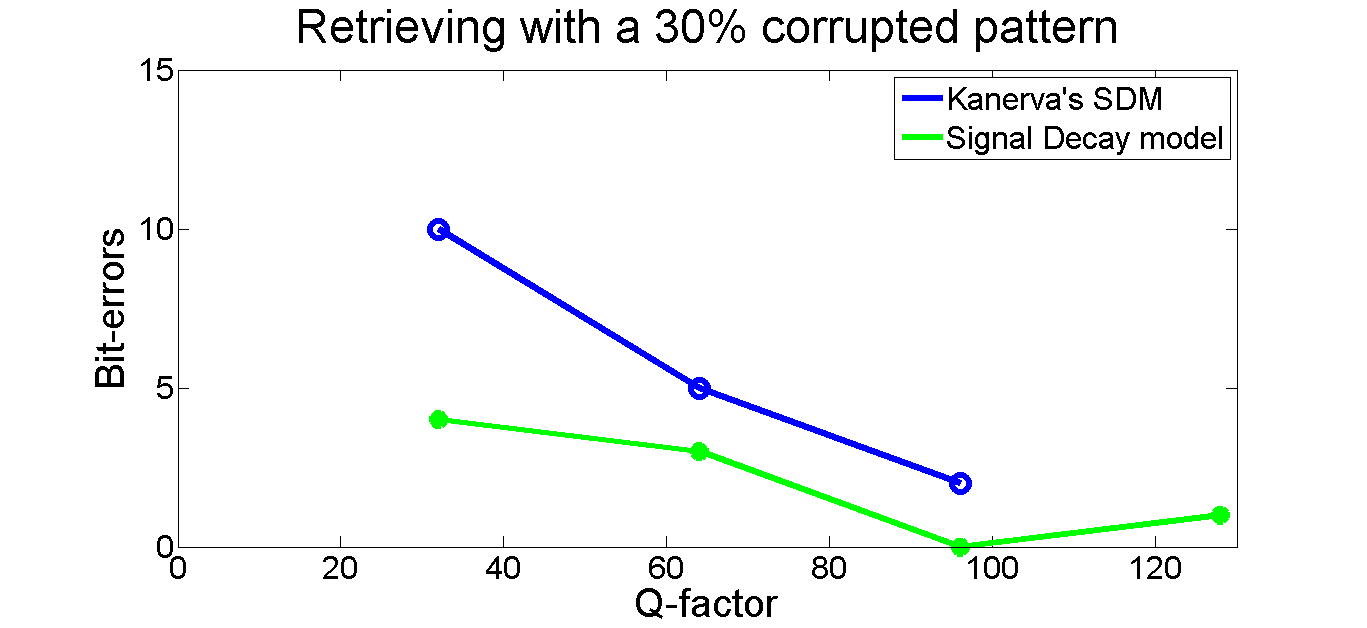}
\captionof{figure}{\small{Results of the test performed on a SDM using a retrieval pattern with a percentage of corruption of $30\%$.}}
\end{Figure}
\begin{Figure}
\label{fig:test3}
\centering
\includegraphics[width=\columnwidth]{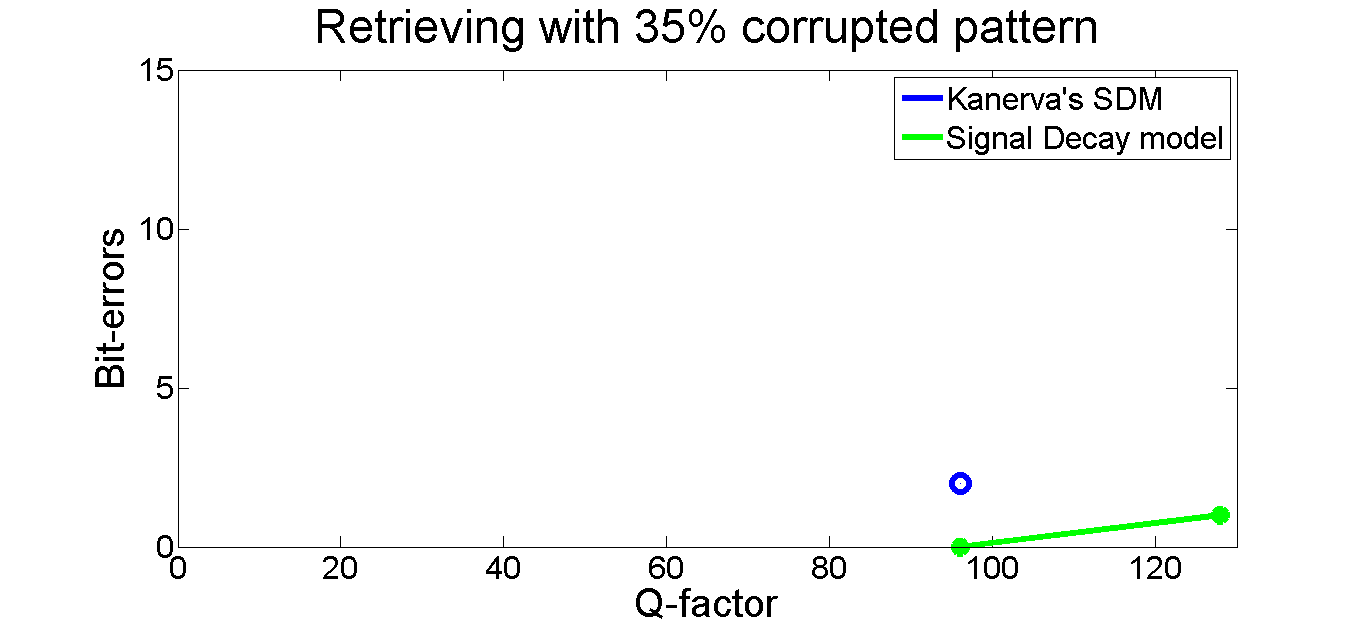}
\captionof{figure}{\small{Results of the test performed on a SDM using a retrieval pattern with a percentage of corruption of $35\%$. Visualizes how the Kanerva's SDM is able to retrieve a pattern only for the one whose Q-factor is $96$.}}
\end{Figure}

Even in this case where the retrieval pattern is corrupted by $35\%$ Kanerva's SDM is still the one which got the worst performances (figure~$7$).
		
\begin{Figure}
\label{fig:test4}
\centering
\includegraphics[width=\columnwidth]{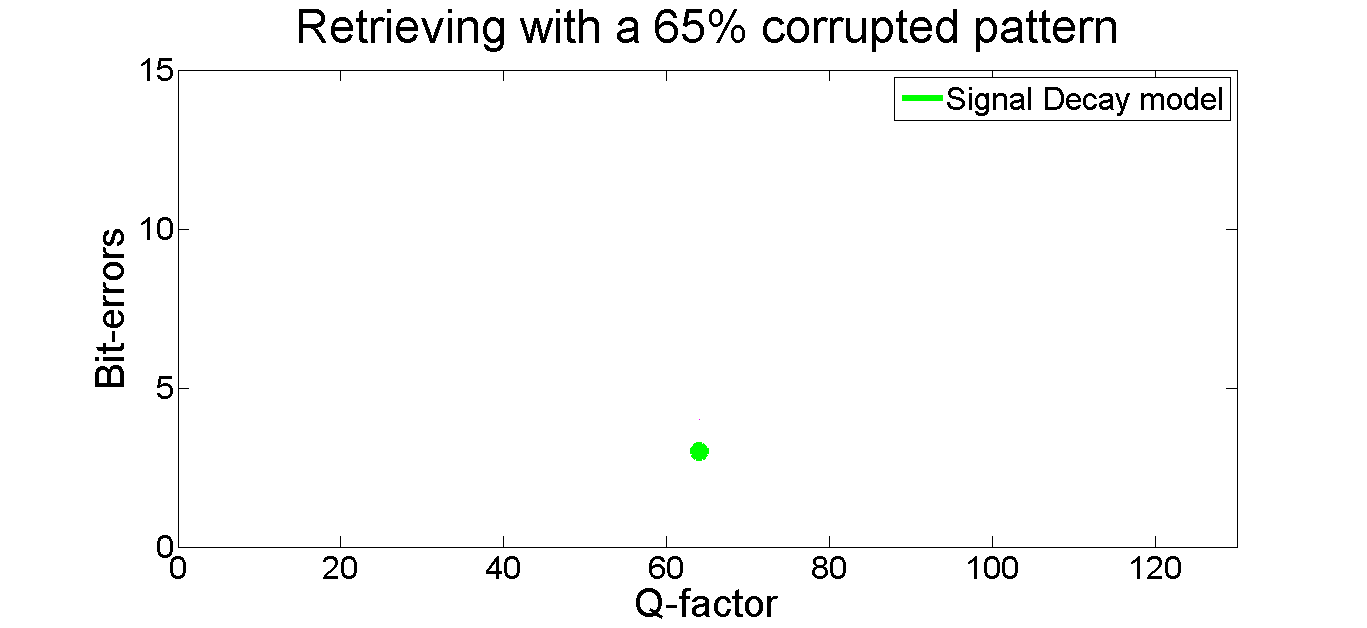}
\captionof{figure}{\small{Results of the test performed on a SDM using a retrieval pattern with a percentage of corruption of $65\%$. With a that corrupted input retrieval pattern only in the case that the Q-factor is $96$ the retrieval ended successfully.}}
\end{Figure}
\begin{Figure}
\label{fig:test5}
\centering
\includegraphics[width=\columnwidth]{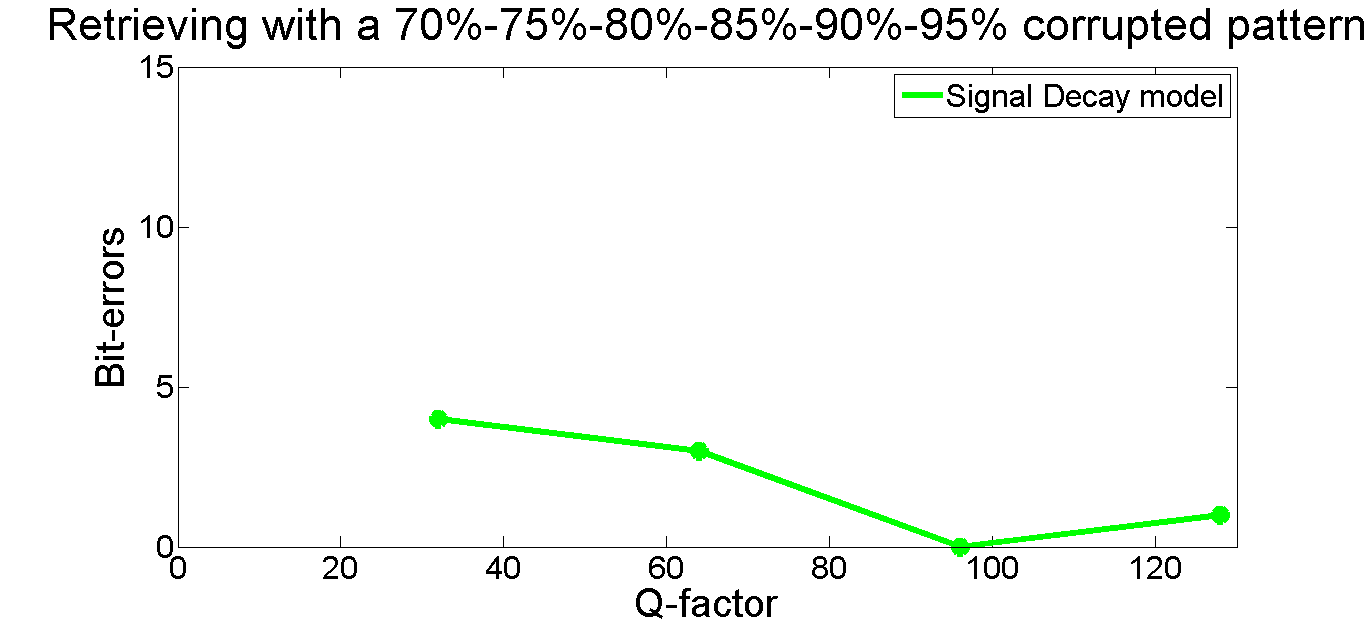}
\captionof{figure}{\small{Results of the test performed on a SDM using a retrieval pattern with a percentage of corruption between $70\%$ and $95\%$.}}
\end{Figure}
		
		It can seem strange that with a Q-factor of $128$ the performances of the SDMs deteriorate a little, but it is important to remember that the patterns used both for the training and for the retrieval are randomly chosen, so that deterioration could be imputated to an unlucky combination of choice.
		
		Going on to analize the results of the tests, in figure~$8$ and~$9$ is possibile to see how the Signal Decay model is able to retrieve patterns corrupted by a percentage of corruption between $65\%$ and $95\%$ committing less than $5$ bit-errors. Obviously only the Signal Decay model is able to retrieve a pattern from a that highly corrupted input.
		
After the accomplishment of the tests it is possible to notice that the new model always have better performances than Kanerva's SDM. In addition the Signal Decay model is able to recognize highly corrupted patterns.

		\subsection{Discussion}
		The purpose of this article is to review Kanerva's SDM model, whose charm comes from Kanerva's idea to get inspiration from biology trying to emulate human's brain behaviour. 
		
		As Denning said in~$[2]$ there are many ways to enhance the model of the Spare Distributed Memory, such as the capability to recognize patterns independently from traslations, rotations or zooms. And I hope that this article could arouse the attention for this problems using the model of the Sparse Distributed Memory.
		
		The Signal Decay model presented in this article overcome the Kanerva's SDM limit of efficently handling only random data, and give the SDM a characteristic inspired by human's brain capacity such as the recognition of inverted pattern. This model is also more flexible and dynamic thanks to the utilization of a hard locations creation process which depend on the input patterns.
		
		The desire is that this article could help to make a further step towards what in the future will hopefully be able to replicate the human memory, and moreover the human recognition capability using the promising Kanerva's theory.
		}
		\end{multicols}
		
		\subsection*{References}
		\small{
		\begin{itemize}
			\item $\left[1\right]$ Kanerva, P. (1993) Associative Neural Memories. \emph{Sparse Distributed Memory and related models}, chapter 3. Oxford University Press.
			\item $\left[2\right]$ Denning, P. J. (1989) Sparse Distributed Memory. \emph{American Scientist 77} (July-August), 333-335.
			\item $\left[3\right]$ Kanerva, P. (1997) Fully Distributed Representation. \emph{In Proceedings RWC Symposium}, 358-365.
			\item $\left[4\right]$ Hely, T. A., Willshaw, D. J. and Hayes, G. M. (1999) A New Approach to Kanerva's Sparse Distributed Memory. \emph{IEEE Transactions on Neural Networks}.
			\item $\left[5\right]$ Rogers, D. (1988) Kanerva's Sparse Distributed Memory: An Associative Memory Algorithm Well-Suited to the Connection Machine, \emph{Technical Report 88.32}. Research Institute for Advanced Computer Science.
			\item $\left[6\right]$ Landi, G. (2011) Elementi di Teoria delle Reti Neurali.
		\end{itemize}
		}
		\end{document}